\documentclass{article} %
\usepackage{iclr2024_conference,times}

\usepackage{amsmath,amsfonts,bm}

\def\eqref#1{equation~\ref{#1}}

\def\1{\bm{1}}

\DeclareMathAlphabet{\mathsfit}{\encodingdefault}{\sfdefault}{m}{sl}
\SetMathAlphabet{\mathsfit}{bold}{\encodingdefault}{\sfdefault}{bx}{n}

\usepackage{hyperref}       %
\usepackage{url}            %
\usepackage{booktabs}       %
\usepackage{amsfonts}       %
\usepackage{nicefrac}       %
\usepackage{microtype}      %
\usepackage{xcolor}         %
\usepackage{graphicx}       %
\usepackage{multirow}       %
\usepackage{caption}
\usepackage{enumitem}
\usepackage{subcaption}

\title{Hypernymy Understanding Evaluation of\\Text-to-Image Models via WordNet Hierarchy}

\author{%
  Anton Baryshnikov
  \thanks{Equal contribution. Correspondence to \texttt{mryabinin0@gmail.com}.}
  \\
  HSE University, Yandex\\
  \texttt{anthony.baryshnikov@gmail.com} \\
  \And
  Max Ryabinin\footnotemark[1]\\
  HSE University, Yandex\\
  \texttt{mryabinin0@gmail.com}
}

\usepackage[capitalise, noabbrev]{cleveref}

\newcommand{\SCS}{\mathrm{SCS}}
\newcommand{\ISP}{\mathrm{ISP}}

\newcommand\outline[1]{\leavevmode%
  \def\maltext{#1}%
  \setbox\qbox=\hbox{\maltext}%
  \boxgs{Q q 2 Tr \thickness\space w 0 0 0 rg 0 G}{}%
  \copy\qbox%
}

\iclrfinalcopy

\begin{document}

\maketitle

\begin{abstract}
  Text-to-image synthesis has recently attracted widespread attention due to rapidly improving quality and numerous practical applications.
  However, the language understanding capabilities of text-to-image models are still poorly understood, which makes it difficult to reason about prompt formulations that a given model would understand well.
  In this work, we measure the capability of popular text-to-image models to understand \textit{hypernymy}, or the ``is-a'' relation between words.
  We design two automatic metrics based on the WordNet semantic hierarchy and existing image classifiers pretrained on ImageNet.
  These metrics both enable broad quantitative comparison of linguistic capabilities for text-to-image models and offer a way of finding fine-grained qualitative differences, such as words that are unknown to models and thus are difficult for them to draw.
  We comprehensively evaluate popular text-to-image models, including GLIDE, Latent Diffusion, and Stable Diffusion, showing how our metrics can provide a better understanding of the individual strengths and weaknesses of these models.
\end{abstract}

\section{Introduction}
\label{sect:intro}

Over the past several years, text-to-image generation has demonstrated remarkable advances~\citep{dalle, glide, ldm, dalle2, imagen} in the quality of generated samples, allowing users to create high-fidelity images from a prompt in natural language.
These improvements have enabled a variety of practical applications, marking a visible shift in the paradigm of conditional image generation.

Despite the progress in this field, the evaluation of images generated from textual input is still a challenging task.
In particular, the majority of works relies on standard metrics for unconditional image generation, such as the Frechet Inception Distance (FID,~\citealp{fid}) on datasets of images paired with their captions, for example, MS-COCO~\citep{coco}.
As this metric uses captions only as model prompts, it provides an implicit measure of language understanding; similarly, caption-to-image similarity using CLIP~\citep{clip} also does not offer a fine-grained way to understand the language comprehension abilities of the network.
However, as correctly visualizing the prompt requires \textit{understanding} the prompt, we are ultimately interested in methods for more in-depth analysis of the model's linguistic competencies.

Several aspects of language understanding are of interest to users of text-to-image generation systems.
For example, one crucial aspect is \textit{knowledge of the meaning} of a term: asking a model to depict an object by giving a word that it has not observed during training is unlikely to be successful.
Also, if a model is able to draw \textit{only one particular subclass} of an object (for example, only one dog breed when asked to draw a dog) across many samples, it significantly restricts the creative potential of the user for a prompt containing such an object.
Even if it is possible to generate an object of another subclass, knowing ``difficult categories'' for a model in advance can reduce the amount of manual effort and help the user find a model more suitable for their goals.

In this work, we build tools for analyzing the \textit{lexical semantics} capabilities in text-to-image generation models.
To construct the metrics for such analysis, we leverage WordNet~\citep{wordnet}, a well-known lexical database of English words annotated with several semantic relations.
Among these relations, we focus on \textit{hypernymy}, or the ``is-a'' relation.
Simply put, hypernymy is the relation between a more general term (for example, ``an animal''), called \textit{a hypernym}, and a more specific term (for example, ``a dog''), called \textit{a hyponym}.

\begin{figure}[t!]
    \centering
    \includegraphics[width=\textwidth]{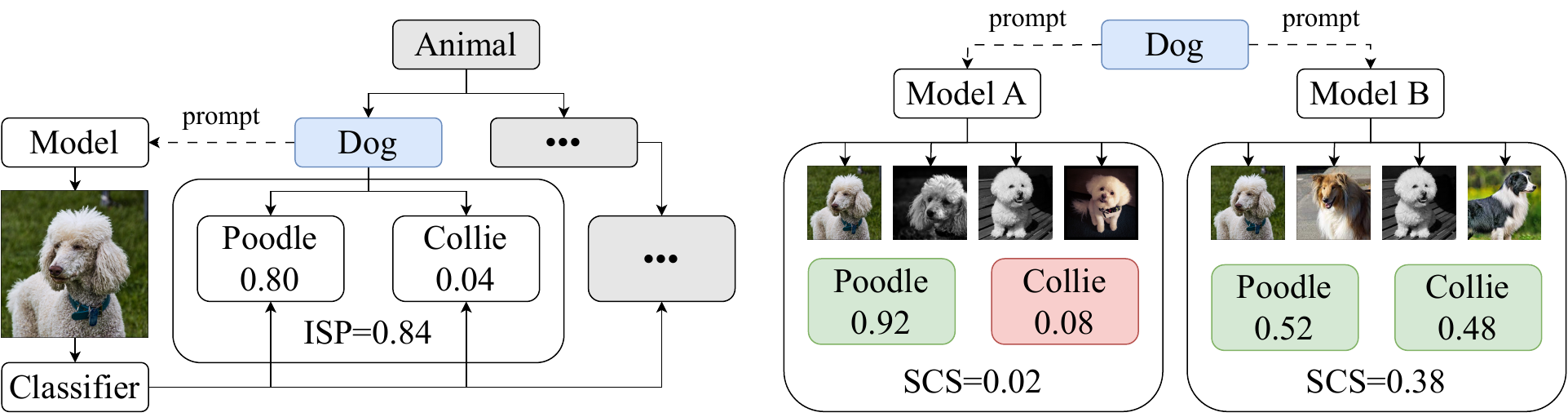}
    \caption{Example computation of the In-Subtree Probability (left) and the Subtree Coverage Score (right). Blue color marks the synset used as a prompt.}
    \label{fig:teaser}
    \vspace{-12pt}
\end{figure}

Using the hypernymy tree from WordNet, we can prompt the model with a specific term (called a \textit{synset}) and measure whether samples of the model with this prompt are in the subtree of the term's hyponyms.
Crucially, the WordNet synsets are a superset of classes of ImageNet~\citep{imagenet}, a highly popular dataset for training image classifiers.
This correspondence allows us to locate the generated samples in the hierarchy using off-the-shelf models pretrained on ImageNet.
More specifically, we design two text-to-image generation metrics for measuring the understanding of semantics.
The first one, named the \textit{In-Subtree Probability} ($\ISP$), shows how well a model generates instances of an object given a specific prompt, while the second one, called the \textit{Subtree Coverage Score (SCS)}, displays the coverage of the hyponym subtree for that prompt.
\cref{fig:teaser} contains an ISP and SCS calculation example for a single synset.

We compute ISP and SCS for several popular models, such as GLIDE~\citep{glide}, Latent Diffusion~\citep{ldm}, Stable Diffusion, and unCLIP~\citep{dalle2}, showing that our metrics generally agree both with existing metrics for text-to-image generation and with human evaluation results.
Importantly, the granular nature of our metrics enables a more detailed analysis of linguistic competencies: for example, we show that it is possible to use ISP to find concepts (or meanings of words) unknown to the model.
In addition, one can use ISP and SCS to easily compare the performance of two models for a particular set of domains or find domains with the highest disparity between models.
We also provide a preliminary analysis of the reasons behind the varying performance of models on different synsets.
As we demonstrate, the capability of a model to generate correct hyponyms is connected with the hypernymy knowledge of its language encoder and the frequency of specific synsets in its training data.

In summary, the main contributions of this paper are as follows:
\begin{itemize}
\item We propose an evaluation framework for text-to-image generation models that leverages the WordNet hierarchy to assess their hypernymy knowledge.
Specifically, we design two interpretable metrics, the In-Subtree Probability and the Subtree Coverage Score, that measure the generation precision and the coverage of the WordNet tree across prompts.
\item We evaluate a broad range of publicly available models, including Latent Diffusion and Stable Diffusion, using the proposed metrics\footnote{The code of our experiments is at \href{https://github.com/yandex-research/text-to-img-hypernymy}{\texttt{github.com/yandex-research/text-to-img-hypernymy}}}. We study the influence of the classifier-free guidance scale~\citep{ho2021classifierfree}, the number of diffusion steps, and the number of generated samples on the behavior of our metrics.
\item We provide an example analysis of language understanding capabilities for popular text-to-image models made possible by our evaluation framework.
Specifically, we show how to use the In-Subtree Probability and the Subtree Coverage Score to find concepts that are less known to the model or less diverse in their hyponym distribution.
\item We study the connection between the text-to-image model performance (according to ISP and SCS) and the occurrence of relevant concepts in its training data. 
We compare the per-synset results of text-to-image models and the frequency of objects in standard datasets for training such models, showing that the correlation is higher for weaker models.
\end{itemize}
\section{Background}
\label{section:background}

\subsection{Text-to-image generation}
Models for generating images from textual prompts have rapidly improved in recent years.
Starting from the release of DALL-E~\citep{dalle} and marked by the emergence of diffusion models~\citep{diffusion-orig,ddpm}, the field has undergone a steady increase in sample fidelity and diversity.
Most popular text-to-image models of today, such as Latent Diffusion (LDM,~\citealp{ldm}), Stable Diffusion (SD,~\citealp{ldm}), and Imagen~\citep{imagen}, rely on sampling from the reverse diffusion process.
The forward diffusion process gradually adds Gaussian noise to images, eventually transforming them into a stationary distribution, and the model learns the reverse process (i.e., generating images from noise) by optimizing a denoising objective. 
The diffusion process can be controlled with several hyperparameters: the number of diffusion steps, the noise schedule (the magnitude of noise added at each step), and the solver type. 
These hyperparameters directly affect the quality of samples: for instance, increasing the number of diffusion steps generally results in higher image fidelity~\citep{progressive_sampling}.

Generating images that would correspond to a given caption is usually done by conditioning diffusion models on the natural language input with a pretrained encoder like BERT~\citep{bert} or the textual encoder of CLIP~\citep{clip}.
It is also possible to trade off caption alignment and sample diversity with classifier-free guidance~\citep{ho2021classifierfree}. 
This technique blends the conditional and unconditional diffusion processes with weights $w$ and $1 - w$, respectively. 
Generally, increasing $w$ results in higher similarity between the caption and the image, while decreasing it results in more diverse images.

\subsection{Quality metrics for text-to-image synthesis}

The standard practice of the research community is to evaluate text-to-image models in terms of sample quality and the similarity of the image to the prompt.
Image quality is usually measured in terms of the Inception Score (IS,~\citealp{is_cite}) and the Fréchet Inception Distance (FID,~\citealp{fid}): the first metric uses the outputs of a pretrained ImageNet classifier to estimate the diversity and fidelity of images, while the second metric computes the similarity between representations of model outputs (also extracted from a pretrained model) and representations of a reference image dataset.
These metrics assess purely visual aspects of model outputs; by contrast, CLIPScore measures the text-image alignment as the cosine similarity between CLIP embeddings of the prompt and the resulting sample.
Although this metric reflects the direct understanding of the prompt, it only measures a surface-level ability to depict that prompt and does not measure the ability of the model to cover the overall visual hierarchy.
Moreover, the lack of a predefined hierarchy makes it difficult to derive a holistic proxy measure of model performance across all object categories.

In addition to the above approaches, there exist metrics that target more nuanced skills of text-to-image generation models.
Namely, Semantic Object Accuracy~\citep{soa} measures the ability of a model to depict several objects in the same image using a pretrained object detector.
\cite{composit} study the ability of text-to-image generators to generalize to novel combinations of objects and their colors or shapes.
PaintSkills~\citep{paintskills} evaluates object recognition, counting, and spatial relation understanding, as well as gender and skin tone biases.
Similarly, TISE~\citep{tise} proposes specific metrics for object fidelity, positional alignment, and counting alignment in text-to-image models.
Our work also targets a specific aspect of text-to-image generation; however, unlike the aforementioned studies, we measure more abstract abilities of language understanding \textit{beyond} strict adherence to the input text.

\subsection{Linguistic capabilities of text-to-image models}

Despite the popularity of models for text-to-image synthesis, the research into their language understanding has mostly been limited to surface-level abilities such as numeracy or compositionality.
One particular line of work~\citep{daras2022discovering, millire2022adversarial, struppek2022exploiting} examines the sensitivity of text-to-image models to morphology and spelling phenomena such as homoglyphs (pairs of similarly looking symbols).
However, to the best of our knowledge, no prior studies have focused on the broader semantic capabilities of such models.
Our work addresses this gap by evaluating both overall awareness of the concept hierarchy and the variety of hyponyms for individual concepts.
\section{Methodology}
\label{section:method}

This section describes our proposed mechanism for measuring the understanding of hypernymy in text-to-image generation.
Specifically, we define the sampling protocol that uses the WordNet database for prompts and introduce two metrics that leverage the structure of WordNet combined with the predictions of ImageNet classifiers for those samples.

\subsection{Obtaining samples using the WordNet tree}

As mentioned in \cref{sect:intro}, we want to design a metric for hypernymy knowledge of text-to-image models.
Hence, we rely on existing annotations for hypernymy in the form of WordNet and map the generated images to nodes in WordNet using pretrained ImageNet classifiers.

However, not all WordNet concepts (grouped into synonym sets or \textit{synsets}) have corresponding classes in the ImageNet dataset, especially in its version with 1,000 classes.
Thus, for each class of ImageNet-1k, we take its corresponding synset in the WordNet hierarchy; we call these synsets \textit{leaf nodes}, and we denote the set of leaf nodes as $L$.
After obtaining $L$, we take all WordNet synsets that are hypernyms of these leaf nodes and use their union as our evaluation set.
For example, for the ImageNet class ``green lizard'', its hypernyms would include nodes such as ``lizard'', ``reptile'', ``organism'', and ``physical entity''.
We call the set of leaf nodes that can be reached from the synset $s$ its \textit{classifiable subtree}, denoted as $\mathrm{A}(s)$. 
Notably, the leaf nodes are excluded from the evaluation set because their classifiable subtrees are empty.

Next, we sample a set of images according to the following protocol: for each concept $s$ in the evaluation set, we take its first lemma name and use it as a prompt for a text-to-image model.
Each lemma is substituted into the template ``An image of {a/an} {lemma}.'' (e.g., ``An image of a dog.'', ``An image of an oven.''); in our preliminary experiments, we found that all templates from the set of prompts recommended by~\cite{clip} yield similar results.
We denote the set of generated images for the synset $s$ as $X_s$. We resize the generated images to 224 $\times$ 224 using bilinear interpolation to match the input dimensions of ImageNet classifiers.

After we generate samples for each concept, we obtain the class probability distribution $p(y|x)$ for each sample $x$ using a pretrained ImageNet classifier.
We then calculate the \textit{hyponym probability distribution} $p_s(y | x)$ for each generated image $x$ of a synset $s$: it is computed as the conditional class distribution given that the generated image is in the classifiable subtree of $s$.
More formally,
\begin{equation}
    p_s(y | x) = p\left(y | x, y \in A(s)\right),
\end{equation}
which can be obtained by taking the $\mathrm{softmax}$ of classifier logits over the subset of classes corresponding to the classifiable subtree of $s$.
We also define the average distribution of hyponyms $\hat{p}_s(y)$ for the synset $s$ as the following expression:
\begin{equation}
    \hat{p}_s(y) = \frac{1}{|X_s|} \sum_{x \in X_s} p_s(y | x).
\end{equation}

Having computed the probability distribution over hyponyms, we can now design two metrics that leverage this distribution to measure different aspects of hyponymy understanding.

\subsection{In-Subtree Probability}
First, we would like to measure the correctness of generation: we expect the model to generate less abstract interpretations of the prompt word (i.e., children nodes according to the WordNet hierarchy) and not to generate unrelated concepts.
The first metric is called the \textbf{In-Subtree Probability (ISP)}: we define it as the probability that the generated image lies in the classifiable subtree of the prompt's synset.
We average the probabilities over generated images for each synset.
The formula for computing ISP is as follows:
\begin{equation}
    \ISP(s) =
    \frac{1}{|X_s|} \sum\limits_{x \in X_s} \sum\limits_{c \in A(s)} p(c | x),
\end{equation}

Naturally, higher values of ISP correspond to outputs that are more consistent with the expectations of the user, and the ideal ISP value is equal to 1.

\subsection{Subtree Coverage Score}
\vspace{-2pt}
\label{scs_subsection}

For the second metric, we want to describe the diversity of generated outputs according to the hypernymy relation.
Intuitively, we are interested in covering the entire subtree of the synset across many samples while ensuring that each sample represents \textit{only one} object.
This prevents two undesirable failure modes: outputs showing ``a mixture'' of many objects and outputs depicting only one hyponym of the concept.
Such properties of unconditional image generators are evaluated by the Inception Score~\citep{inceptionscore}, which is why we follow it in the design of our metric named the \textbf{Subtree Coverage Score (SCS)}. 
For each concept $s$, we calculate the average Kullback-Leibler divergence between the hyponym probability distribution and the average distribution of hyponyms across all samples generated from $s$ as a prompt:
\begin{equation}
    \SCS(s) = \frac{1}{|X_s|} \sum\limits_{x \in X_s} \mathrm{D_{KL}}(p_s(y | x) | \hat{p}_s(y)).
\end{equation}
As with the Inception Score and ISP, the higher the value of SCS, the better.

\subsection{Aggregating results}
Each of the above metrics measures the results for a single synset.
To get the final metric value for a single model, we average the metrics across all synsets from the evaluation set and divide the result by the maximum possible value (1.0 for ISP and $\approx$1.624 for SCS) for ease of interpretation. 
One may also note that $\SCS(s)$ is always equal to 0 when $s$ has only one node in $\mathrm{A}(s)$, as it reduces to the average of Kullback-Leibler divergences for identical distributions.
Therefore, we exclude these synsets from aggregation in the case of the Subtree Coverage Score; however, we keep them when calculating the model's In-Subtree Probability.

This direct averaging treats all synsets equally regardless of their position in the WordNet hierarchy, causing the metrics to be incomparable between synsets from different levels.
Indeed, higher nodes have more hyponyms by construction: for instance, the value of ISP for ``entity'' (the root of the WordNet tree) is always equal to 1.
As a result, values from different levels of WordNet might skew the aggregated metric.
Future work might address this issue by applying a discounting factor to higher levels of the hierarchy.
However, in this paper, we aim to introduce the approach of hierarchical evaluation and thus leave this question out of the scope of our study.

\section{Experiments}
\label{section:experiments}

In this section, we evaluate several popular text-to-image models with ISP and SCS to compare our metrics with other approaches, including human evaluation.
We also study the influence of several generation hyperparameters on the behavior of the proposed metrics.

\subsection{Setup}
\label{section:setup}

\nocite{pytorch,numpy,plt,diffusers,torchvision2016}

We run the experiments on the following text-to-image models: GLIDE~\citep{glide}, Latent Diffusion~\citep{ldm}, Stable Diffusion 1.4, Stable Diffusion 2.0, and unCLIP~\citep{dalle2}.
We use an open-source version of unCLIP~\citep{karlo} in our experiments, as the original one is not publicly available.
We chose these models because they are openly available and were close to state-of-the-art at the time of their release.
We use ViT-B/16~\citep{vit} as the ImageNet classifier due to its high accuracy and low calibration error.
After experimenting with different pretrained classifiers, we found that they resulted in highly similar rankings for both metrics; more details on the choice of the classifier are available in \cref{appendix:classifier_comparison}.

We generate 32 images for each synset using the default DDIM sampler with $\eta=0$: experiments with other numbers of samples can be seen in \cref{appendix:metric_stability}. 
We run each model in 16-bit precision to speed up the generation process.
We use 50 base model steps with 27 upsampler steps for GLIDE, 50 diffusion steps for Latent Diffusion and all Stable Diffusion models, and 25 prior, 25 decoder and 7 super-resolution steps for unCLIP: \cref{section:num_steps} describes our experiments with other numbers of steps. 
Unless stated otherwise, we set the classifier-free guidance~\citep{ho2021classifierfree} weight to 7.5 in all our experiments.

\subsection{Results}
\label{sect:results}
First, we compare the models using the metrics proposed in \cref{section:method}, along with Fréchet Inception Distance~\citep{fid} and CLIPScore~\citep{hessel-etal-2021-clipscore} as baselines.
This comparison is intended to be a form of a ``sanity check'' for ISP and SCS: one would expect that models generally viewed as better generators would also be better at hypernymy knowledge.
FID and CLIP are computed on 10,000 random prompts from the MS-COCO~\citep{coco} validation set. 
We present the results of the experiment in \Cref{tab:main_metrics}: notably, the ranking of models is mostly consistent within metrics of similar categories. 
Both ISP and SCS have a relative standard deviation of less than $1\%$ when computed over four random seeds.

\begin{table}
    \centering
    \caption{Model performance measured by ISP, WIS and baseline metrics. The best values are in bold.}
    \label{tab:main_metrics}
    \begin{tabular}{lcccc}
        \toprule
        \multirow{2}{*}{Model} & \multicolumn{2}{c}{Precision} & \multicolumn{2}{c}{Diversity}\\
        \cmidrule(l){2-3} \cmidrule(l){4-5}
        & ISP $\uparrow$ & CLIPScore $\uparrow$ & SCS $\uparrow$ & FID $\downarrow$ \\\midrule

        GLIDE & 0.221 & 0.279 & 0.198 & 37.93  \\
        Latent Diffusion & 0.217 & 0.304 & 0.182 & 36.43  \\
        Stable Diffusion 1.4 & 0.329 & 0.314 & \textbf{0.256} & 16.57  \\
        Stable Diffusion 2.0 & 0.297 & 0.317 & 0.233 & \textbf{16.25} \\
        unCLIP & \textbf{0.352} & \textbf{0.322} & 0.194 & 18.29  \\
        \bottomrule
    \end{tabular}
\end{table}

\subsection{Human evaluation}
\label{section:human_eval}

In this experiment, we measure the correlation of the In-Subtree Probability and the Subtree Coverage Score with the human understanding of hyponymy.
To do this, we conduct crowdsourced evaluations of text-caption similarity and sample diversity for several text-to-image models.
To estimate text-to-caption similarity, we present the annotators with two generated images along with the caption from which they were generated.
The workers are then tasked to select the image that best matches the text description.
For the diversity evaluation, we show the annotators two collections of generated images and ask them to select the grid with more diverse samples.

The models are evaluated on a random subset of 20 synsets from the WordNet hierarchy; we generate 20 pairs of images (or grids) per concept, which results in 400 tasks per comparison with the overlap of 5 labelers.
We also report Krippendorff's $\alpha$~\citep{krippendorff2018content} as a measure of inter-annotator agreement.
Further details of the human evaluation protocol, including the annotation interface, are shown in \Cref{appendix:human_eval}.

\begin{table}[b]
    \vspace{-18pt}
    \centering
    \caption{Results of human preference evaluation for models compared with Stable Diffusion 1.4. Krippendorff's $\alpha$ is given in subscript.}
    \label{tab:human_eval}
    \begin{tabular}{lcccccc}
        \toprule
        \multirow{2}{*}{Model} & \multicolumn{3}{c}{Caption similarity} & \multicolumn{3}{c}{Sample diversity} \\
        \cmidrule(lr){2-4} \cmidrule(lr){5-7} & Human $\uparrow$ &  $\Delta$ISP $\uparrow$ & $\Delta$CLIPScore $\uparrow$ & Human $\uparrow$ & $\Delta$SCS $\uparrow$ & $\Delta$FID $\downarrow$ \\
        \midrule
        Latent Diffusion & 17.1\% \textsubscript{0.75} & -0.112\phantom{-} & -0.010\phantom{-} & 21.9\% \textsubscript{0.58} & -0.074\phantom{-} & 19.86 \\
        unCLIP & 49.1\% \textsubscript{0.82} & 0.023 & 0.080 & 26.8\% \textsubscript{0.63} & -0.062\phantom{-} & 1.72 \\
        SD 1.4 ($w=2.5$) & 25.3\% \textsubscript{0.81} & -0.060\phantom{-} & -0.080\phantom{-} & 57.5\% \textsubscript{0.58} & 0.033 & -5.09\phantom{-} \\
        \bottomrule
    \end{tabular}
\end{table}

We compare Stable Diffusion 1.4 with classifier-free guidance of 7.5 against Latent Diffusion, unCLIP, and Stable Diffusion 1.4 that has a lower guidance value of 2.5.
The results of this evaluation can be seen in \cref{tab:human_eval}: in general, the differences in all metrics follow human preferences.

Next, we compute rank correlations between synset metric differences and annotator preferences to measure detailed agreement, showing the outcome in \cref{tab:synset_human_eval}.
Unlike CLIPScore and the Inception Score (used here due to a lack of references for FID), both ISP and SCS have a moderate yet statistically significant correlation with human preference and thus are better for granular evaluation.

\begin{table}
    \centering
    \caption{Synset-level Spearman rank correlations of metric differences and human preferences. The subscript shows p-values for correlations. The best values in each category are in bold.}
    \label{tab:synset_human_eval}
    \begin{tabular}{lcccc}
        \\\toprule
        \multirow{2}{*}{Model} & \multicolumn{2}{c}{Caption similarity} & \multicolumn{2}{c}{Sample diversity}\\
        \cmidrule(lr){2-3} \cmidrule(lr){4-5}
         & ISP $\uparrow$ & CLIPScore $\uparrow$ & SCS $\uparrow$ & Inception Score $\uparrow$ \\
        \midrule
        Latent Diffusion & \textbf{0.41} \textsubscript{0.00} & -0.63 \textsubscript{0.00} & \textbf{0.52} \textsubscript{0.00} & 0.33 \textsubscript{0.04}\\
        unCLIP & \textbf{0.63} \textsubscript{0.00} & -0.10 \textsubscript{0.53} & \textbf{0.44} \textsubscript{0.00} & 0.38 \textsubscript{0.02} \\
        SD ($w = 2.5$) & \textbf{0.63} \textsubscript{0.00} & \phantom{-}0.59 \textsubscript{0.00} & \textbf{0.40} \textsubscript{0.01} & 0.38 \textsubscript{0.02} \\
        \bottomrule
    \end{tabular}
    \vspace{-12pt}
\end{table}

\subsection{Impact of the number of diffusion steps}
\label{section:num_steps}

When evaluating machine learning models, one needs to balance the metric computation time and the measurement accuracy.
In the case of diffusion models, this can be easily done by adjusting the number of steps in the reverse diffusion process: fewer steps generally lead to lower image quality.
In this experiment, we aim to determine the optimal number of steps that would be necessary for ISP and SCS.
Specifically, we compute these two metrics on Latent Diffusion and Stable Diffusion 1.4 with the number of steps $T$ from the following set: \{5, 10, 15, 25, 50, 75, 100\}.

\begin{figure}[h]
    \centering
    \includegraphics[width=\textwidth]{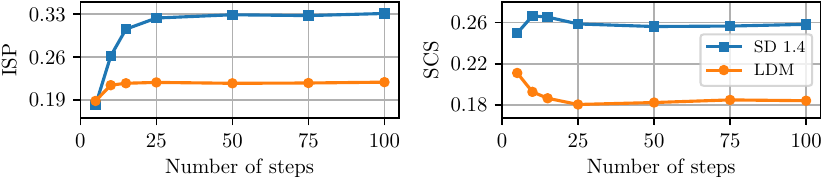}
    \vspace{-12pt}
    \caption{ISP and SCS values depending on the number of diffusion steps.}
    \label{fig:num_steps_sweep}
    \vspace{-4pt}
\end{figure}

\cref{fig:num_steps_sweep} displays the outcome of this experiment: we find that both ISP and SCS are unstable when the number of steps is less than 25, which is expected because the quality of images deteriorates when $T$ is too low~\citep{progressive_sampling}. 
However, increasing the number of diffusion steps beyond this point has little to no effect on the results.
We also note that, unlike the In-Subtree Probability, the Subtree Coverage Score increases at small values of $T$. 
We attribute this to the fact that SCS measures the diversity of classifier predictions, which might be high for out-of-distribution inputs or images with excessive noise.

\vspace{-4pt}
\subsection{Impact of classifier-free guidance}
As we discussed in~\cref{section:background}, classifier-free guidance is a technique that allows trading off sample precision for diversity.
To study the influence of the guidance weight on our metrics, we repeat the experiments of \cref{sect:results} for all models using the $w$ values of \{1.0, 1.5, 2.0, 2.5, 5.0, 7.5, 10.0\}.

\begin{figure}[h]
    \vspace{-4pt}
    \centering
    \includegraphics{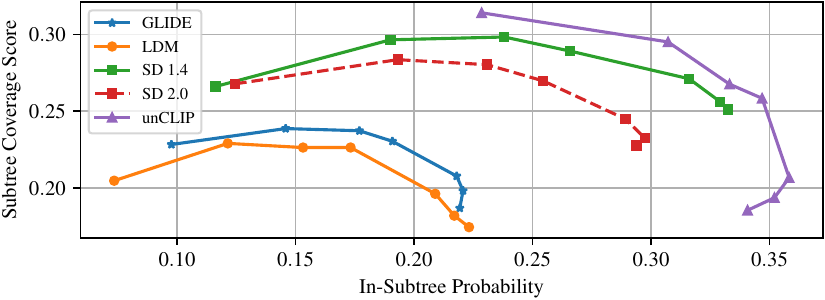}
    \vspace{-14pt}
    \caption{Results of evaluation with different guidance scales.}
    \label{fig:guidance_sweep}
\end{figure}

Our findings are shown in \cref{fig:guidance_sweep}: as anticipated, higher guidance leads to better precision (indicated by higher ISP), and lower guidance leads to more diverse samples (as indicated by higher SCS).
We note that excessively high or low guidance values may result in both lower SCS and lower ISP, which hints at the presence of generation artifacts. 
We also observe that the relative positions of Pareto frontiers rarely intersect even for extreme guidance values: this means that it is possible to use our metrics with different guidance scales depending on the application and expect similar results.
Intuitively, hypernymy knowledge is a skill that is independent of high-fidelity image generation ability, which is consistent with the results we obtain here.
\section{Analysis}

\subsection{Finding unknown concepts}
\label{section:low_isp}
\begin{figure}[h]
    \centering
    \includegraphics[width=\textwidth]{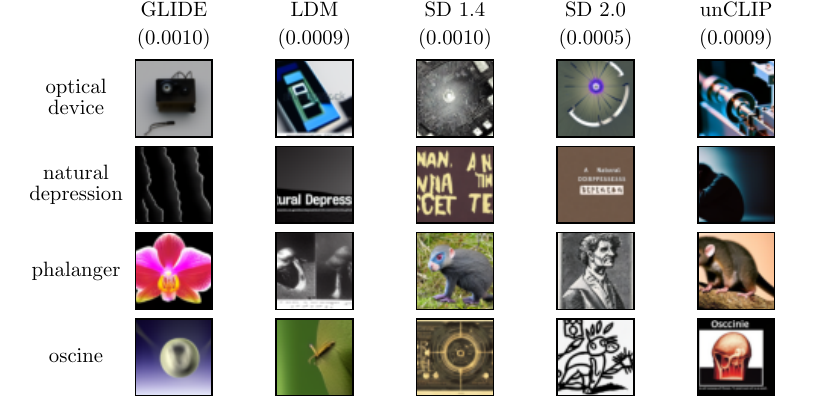}
    \vspace{-4pt}
    \caption{Outputs for synsets with low In-Subtree Probability. Average model ISP for these synsets is shown in parentheses.}
    \label{fig:low_in_synsets}
    \vspace{-6pt}
\end{figure}

Using the In-Subtree Probability, we can easily determine which concepts are drawn poorly by the model by taking synsets with low values of this metric. 
To demonstrate this use case, we select synsets that are among the lowest ones in terms of ISP across different models.
In \Cref{fig:low_in_synsets}, we show a subset of these concepts chosen for illustration purposes, and a random selection of synsets is presented in \Cref{appendix:low_in_random}. 

As we can see, our approach not only uncovers inherently unknown concepts (such as ``phalanger'' or ``oscine''), but also detects homonyms for which the models are only familiar with one meaning (e.g., ``convertible'' or ``landing'').
Additionally, it identifies synsets where the models only recognize some of its hyponyms (e.g., ``contestant''). 
In some cases, the model generates a coherent output, but the concept understanding is insufficient to achieve high ISP (e.g., ``optical device'').
We perform an identical analysis with the Subtree Coverage Score to find concepts with low diversity in \Cref{appendix:low_diversity}.

\subsection{Granular comparison of models}

\begin{figure}[h!]
    \vspace{-6pt}
    \centering
    \includegraphics{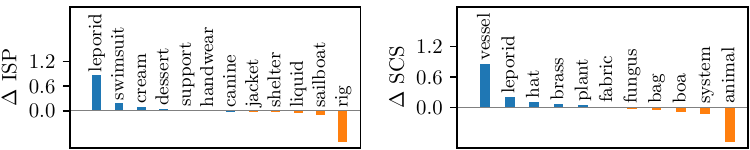}
    \caption{Per-synset comparison between Stable Diffusion 1.4 and Stable Diffusion 2.0. The vertical axis denotes the differences in metrics between the former and the latter model.}
    \label{fig:granular_comp}
    \vspace{-10pt}
\end{figure}

We can also use our metrics to compare two models in terms of how well they generate individual concepts.
To do this, we calculate the differences between ISP and SCS for each synset and rank synsets according to the resulting differences.
We present this analysis for Stable Diffusion 1.4 and Stable Diffusion 2.0 in Figure \ref{fig:granular_comp}.
Such comparison allows us to more easily understand each model's relative strengths and weaknesses with direct illustrations.
For instance, we can see that the models are almost always equal in performance and yet still have synsets with drastic metric differences. 
Apart from analyzing model performance on specific concepts, it is also possible to evaluate them on entire synset subtrees, which we show in \Cref{appendix:subtree_comp}.

\subsection{Relationship with training data}
We hypothesize that poor representation of some concepts may depend on their frequency in the training corpus. 
We analyze three popular multimodal datasets: LAION-400M~\citep{laion400m}, LAION-2B-en~\citep{laion5b} and COYO~\citep{coyo}, counting the number of times that each WordNet concept appeared in the text captions.
These three datasets have a significant presence in the training data of the models we use: Latent Diffusion was trained on LAION-400M, Stable Diffusion 1.4 was trained on LAION-2B-en and then finetuned, Stable Diffusion 2.0 was trained on a superset of LAION-2B-en and then finetuned, and the unCLIP variation we used was partially trained on COYO.
After computing the frequencies, we measure the Spearman rank correlation between the synset counts and the per-synset metrics of the models we evaluate in our primary experiments.

\begin{table}[h]
    \centering
    \vspace{-2pt}
    \caption{Spearman rank correlation between synset metrics and their frequency in the dataset. $p$-values are in subscript, statistically significant results ($p <$ 0.05) are in bold.}
    \vspace{-6pt}
    \begin{tabular}{lcccccc}
        \\\toprule
        \multirow{2}{*}{Model} & \multicolumn{3}{c}{In-Subtree Probability} & \multicolumn{3}{c}{Subtree Coverage Score} \\
        \cmidrule(lr){2-4} \cmidrule(lr){5-7}
        & LAION-400M & LAION-2B & COYO & LAION-400M & LAION-2B & COYO \\
        \midrule
        GLIDE & \textbf{0.19}\textsubscript{ 0.00} & \textbf{0.18}\textsubscript{ 0.00} & \textbf{0.16}\textsubscript{ 0.00} & \textbf{0.28}\textsubscript{ 0.00} & \textbf{0.29}\textsubscript{ 0.00} & \textbf{0.29}\textsubscript{ 0.00}\\
        LDM & \textbf{0.29}\textsubscript{ 0.00} & \textbf{0.27}\textsubscript{ 0.00} & \textbf{0.24}\textsubscript{ 0.00} & \textbf{0.15}\textsubscript{ 0.00} & \textbf{0.16}\textsubscript{ 0.00} & \textbf{0.17}\textsubscript{ 0.00}\\
        SD 1.4 & 0.06\textsubscript{ 0.15} & 0.04\textsubscript{ 0.34} & 0.01\textsubscript{ 0.81} & 0.00\textsubscript{ 0.15} & 0.01\textsubscript{ 0.34} & 0.03\textsubscript{ 0.81}\\
        SD 2.0 & \textbf{0.10}\textsubscript{ 0.01} & \textbf{0.08}\textsubscript{ 0.04} & 0.05\textsubscript{ 0.18} & \textbf{0.07}\textsubscript{ 0.01} & \textbf{0.08}\textsubscript{ 0.04} & 0.08\textsubscript{ 0.18}\\
        unCLIP & 0.02\textsubscript{ 0.63} & 0.00\textsubscript{ 0.91} & -0.02\textsubscript{ 0.61}\phantom{-} & 0.04\textsubscript{ 0.63} & 0.05\textsubscript{ 0.91} & 0.08\textsubscript{ 0.61}\\
        \bottomrule
    \end{tabular}
    \label{tab:train_data}
\end{table}

As we can see from \Cref{tab:train_data}, the majority of correlations are not high in magnitude yet still significant, which suggests that hypernymy understanding and concept knowledge cannot be attributed purely to the frequency of specific synsets in training data.
Weaker models also tend to have higher correlations, whereas the results for stronger models are less pronounced. 
This difference might arise due to the finetuning procedures on aesthetic images or simply the higher capacity of better models. 
Alternatively, the hyponymy performance of text-to-image models might arise purely from the semantic capabilities of the part of the model that encodes the prompt. 
In \cref{appendix:textual_encoder}, we provide the results of evaluation for the CLIP language encoder, showing that there is a high and significant correlation between average hyponym embedding similarities and metric values for a given synset.

\vspace{-4pt}
\section{Conclusion}

In this work, we introduce In-Subtree Probability and Subtree Coverage Score, two metrics for evaluating the language understanding capabilities of text-to-image models. We validate these metrics by comparing them to standard evaluation methods and human judgment.
Through extensive analysis, we demonstrate how ISP and SCS can provide a deeper understanding of text-to-image models and their semantic abilities.

Future work might address the limitation of our approach connected to its reliance on WordNet and ImageNet: these datasets do not contain the entire concept hierarchy, and therefore it might be valuable to study data-driven hierarchies (such as the ones proposed by \citealp{meru}) based on the actual use cases of text-to-image models.
Furthermore, models that explicitly leverage ImageNet data (such as all models using the CLIP encoder) might have an unfair advantage due to a smaller domain shift and thus obtain inflated ISP and SCS scores.

\section*{Ethics Statement}

Text-to-image models trained on large-scale web data can generate sensitive or offensive content. 
We do not directly improve these capabilities of the models; instead, we offer a way to more thoroughly monitor their performance, which could help decrease undesired behavior. 
We use human evaluation in our study. 
The workers were paid above the minimum wage in their respective countries; for details, see \Cref{appendix:human_eval}.

\section*{Reproducibility Statement}

Our work makes the following efforts to ensure reproducibility: we release the code for our experiments and analysis, we report the setup of our experiments and hyperparameter choices in \Cref{section:setup}, and we provide details on the human evaluation protocol in \Cref{section:human_eval,appendix:human_eval}.

\bibliographystyle{iclr2024_conference}
\bibliography{bibliography}

\begin{thebibliography}{41}
\providecommand{\natexlab}[1]{#1}
\providecommand{\url}[1]{\texttt{#1}}
\expandafter\ifx\csname urlstyle\endcsname\relax
  \providecommand{\doi}[1]{doi: #1}\else
  \providecommand{\doi}{doi: \begingroup \urlstyle{rm}\Url}\fi

\bibitem[Byeon et~al.(2022)Byeon, Park, Kim, Lee, Baek, and Kim]{coyo}
Minwoo Byeon, Beomhee Park, Haecheon Kim, Sungjun Lee, Woonhyuk Baek, and Saehoon Kim.
\newblock Coyo-700m: Image-text pair dataset.
\newblock \url{https://github.com/kakaobrain/coyo-dataset}, 2022.

\bibitem[Cho et~al.(2022)Cho, Zala, and Bansal]{paintskills}
Jaemin Cho, Abhay Zala, and Mohit Bansal.
\newblock Dall-eval: Probing the reasoning skills and social biases of text-to-image generative transformers.
\newblock \emph{ArXiv}, 2022.

\bibitem[Daras \& Dimakis(2022)Daras and Dimakis]{daras2022discovering}
Giannis Daras and Alexandros~G. Dimakis.
\newblock Discovering the hidden vocabulary of dalle-2, 2022.

\bibitem[Deng et~al.(2009)Deng, Dong, Socher, Li, Li, and Fei-Fei]{imagenet}
Jia Deng, Wei Dong, Richard Socher, Li-Jia Li, Kai Li, and Li~Fei-Fei.
\newblock Imagenet: A large-scale hierarchical image database.
\newblock In \emph{2009 IEEE conference on computer vision and pattern recognition}, pp.\  248--255. Ieee, 2009.

\bibitem[Desai et~al.(2023)Desai, Nickel, Rajpurohit, Johnson, and Vedantam]{meru}
Karan Desai, Maximilian Nickel, Tanmay Rajpurohit, Justin Johnson, and Shanmukha~Ramakrishna Vedantam.
\newblock Hyperbolic image-text representations.
\newblock In \emph{International Conference on Machine Learning}, pp.\  7694--7731. PMLR, 2023.

\bibitem[Devlin et~al.(2018)Devlin, Chang, Lee, and Toutanova]{bert}
Jacob Devlin, Ming-Wei Chang, Kenton Lee, and Kristina Toutanova.
\newblock Bert: Pre-training of deep bidirectional transformers for language understanding.
\newblock \emph{arXiv preprint arXiv:1810.04805}, 2018.

\bibitem[Dinh et~al.(2022)Dinh, Nguyen, and Hua]{tise}
Tan~M. Dinh, Rang Nguyen, and Binh-Son Hua.
\newblock Tise: Bag of metrics for text-to-image synthesis evaluation.
\newblock In \emph{Proceedings of the European Conference on Computer Vision (ECCV)}, 2022.

\bibitem[Dosovitskiy et~al.(2020)Dosovitskiy, Beyer, Kolesnikov, Weissenborn, Zhai, Unterthiner, Dehghani, Minderer, Heigold, Gelly, et~al.]{vit}
Alexey Dosovitskiy, Lucas Beyer, Alexander Kolesnikov, Dirk Weissenborn, Xiaohua Zhai, Thomas Unterthiner, Mostafa Dehghani, Matthias Minderer, Georg Heigold, Sylvain Gelly, et~al.
\newblock An image is worth 16x16 words: Transformers for image recognition at scale.
\newblock \emph{arXiv preprint arXiv:2010.11929}, 2020.

\bibitem[Fellbaum(1998)]{wordnet}
Christiane Fellbaum.
\newblock \emph{WordNet: An Electronic Lexical Database}.
\newblock Bradford Books, 1998.
\newblock URL \url{https://mitpress.mit.edu/9780262561167/}.

\bibitem[Harris et~al.(2020)Harris, Millman, van~der Walt, Gommers, Virtanen, Cournapeau, Wieser, Taylor, Berg, Smith, Kern, Picus, Hoyer, van Kerkwijk, Brett, Haldane, del R{\'{i}}o, Wiebe, Peterson, G{\'{e}}rard-Marchant, Sheppard, Reddy, Weckesser, Abbasi, Gohlke, and Oliphant]{numpy}
Charles~R. Harris, K.~Jarrod Millman, St{\'{e}}fan~J. van~der Walt, Ralf Gommers, Pauli Virtanen, David Cournapeau, Eric Wieser, Julian Taylor, Sebastian Berg, Nathaniel~J. Smith, Robert Kern, Matti Picus, Stephan Hoyer, Marten~H. van Kerkwijk, Matthew Brett, Allan Haldane, Jaime~Fern{\'{a}}ndez del R{\'{i}}o, Mark Wiebe, Pearu Peterson, Pierre G{\'{e}}rard-Marchant, Kevin Sheppard, Tyler Reddy, Warren Weckesser, Hameer Abbasi, Christoph Gohlke, and Travis~E. Oliphant.
\newblock Array programming with {NumPy}.
\newblock \emph{Nature}, 585\penalty0 (7825):\penalty0 357--362, September 2020.
\newblock \doi{10.1038/s41586-020-2649-2}.
\newblock URL \url{https://doi.org/10.1038/s41586-020-2649-2}.

\bibitem[He et~al.(2016)He, Zhang, Ren, and Sun]{resnet}
Kaiming He, Xiangyu Zhang, Shaoqing Ren, and Jian Sun.
\newblock Deep residual learning for image recognition.
\newblock In \emph{Proceedings of the IEEE conference on computer vision and pattern recognition}, pp.\  770--778, 2016.

\bibitem[Hessel et~al.(2021)Hessel, Holtzman, Forbes, Le~Bras, and Choi]{hessel-etal-2021-clipscore}
Jack Hessel, Ari Holtzman, Maxwell Forbes, Ronan Le~Bras, and Yejin Choi.
\newblock {CLIPS}core: A reference-free evaluation metric for image captioning.
\newblock In \emph{Proceedings of the 2021 Conference on Empirical Methods in Natural Language Processing}, pp.\  7514--7528, Online and Punta Cana, Dominican Republic, November 2021. Association for Computational Linguistics.
\newblock \doi{10.18653/v1/2021.emnlp-main.595}.
\newblock URL \url{https://aclanthology.org/2021.emnlp-main.595}.

\bibitem[Heusel et~al.(2017)Heusel, Ramsauer, Unterthiner, Nessler, and Hochreiter]{fid}
Martin Heusel, Hubert Ramsauer, Thomas Unterthiner, Bernhard Nessler, and Sepp Hochreiter.
\newblock Gans trained by a two time-scale update rule converge to a local nash equilibrium.
\newblock In I.~Guyon, U.~Von Luxburg, S.~Bengio, H.~Wallach, R.~Fergus, S.~Vishwanathan, and R.~Garnett (eds.), \emph{Advances in Neural Information Processing Systems}, volume~30. Curran Associates, Inc., 2017.
\newblock URL \url{https://proceedings.neurips.cc/paper_files/paper/2017/file/8a1d694707eb0fefe65871369074926d-Paper.pdf}.

\bibitem[Hinz et~al.(2019)Hinz, Heinrich, and Wermter]{soa}
Tobias Hinz, Stefan Heinrich, and Stefan Wermter.
\newblock Semantic object accuracy for generative text-to-image synthesis.
\newblock \emph{IEEE Transactions on Pattern Analysis and Machine Intelligence}, 44:\penalty0 1552--1565, 2019.
\newblock URL \url{https://api.semanticscholar.org/CorpusID:204949374}.

\bibitem[Ho \& Salimans(2021)Ho and Salimans]{ho2021classifierfree}
Jonathan Ho and Tim Salimans.
\newblock Classifier-free diffusion guidance.
\newblock In \emph{NeurIPS 2021 Workshop on Deep Generative Models and Downstream Applications}, 2021.
\newblock URL \url{https://openreview.net/forum?id=qw8AKxfYbI}.

\bibitem[Ho et~al.(2020)Ho, Jain, and Abbeel]{ddpm}
Jonathan Ho, Ajay Jain, and Pieter Abbeel.
\newblock Denoising diffusion probabilistic models.
\newblock \emph{Advances in Neural Information Processing Systems}, 33:\penalty0 6840--6851, 2020.

\bibitem[Hunter(2007)]{plt}
J.~D. Hunter.
\newblock Matplotlib: A 2d graphics environment.
\newblock \emph{Computing in Science \& Engineering}, 9\penalty0 (3):\penalty0 90--95, 2007.
\newblock \doi{10.1109/MCSE.2007.55}.

\bibitem[Krippendorff(2018)]{krippendorff2018content}
Klaus Krippendorff.
\newblock \emph{Content analysis: An introduction to its methodology}.
\newblock Sage publications, 2018.

\bibitem[Lee et~al.(2022)Lee, Kim, Choi, Kim, Byeon, Baek, and Kim]{karlo}
Donghoon Lee, Jiseob Kim, Jisu Choi, Jongmin Kim, Minwoo Byeon, Woonhyuk Baek, and Saehoon Kim.
\newblock Karlo-v1.0.alpha on coyo-100m and cc15m.
\newblock \url{https://github.com/kakaobrain/karlo}, 2022.

\bibitem[Lin et~al.(2014)Lin, Maire, Belongie, Hays, Perona, Ramanan, Doll{\'a}r, and Zitnick]{coco}
Tsung-Yi Lin, Michael Maire, Serge Belongie, James Hays, Pietro Perona, Deva Ramanan, Piotr Doll{\'a}r, and C.~Lawrence Zitnick.
\newblock Microsoft coco: Common objects in context.
\newblock In David Fleet, Tomas Pajdla, Bernt Schiele, and Tinne Tuytelaars (eds.), \emph{Computer Vision -- ECCV 2014}, pp.\  740--755, Cham, 2014. Springer International Publishing.
\newblock ISBN 978-3-319-10602-1.

\bibitem[Liu et~al.(2022)Liu, Mao, Wu, Feichtenhofer, Darrell, and Xie]{convnext}
Zhuang Liu, Hanzi Mao, Chao-Yuan Wu, Christoph Feichtenhofer, Trevor Darrell, and Saining Xie.
\newblock A convnet for the 2020s.
\newblock In \emph{Proceedings of the IEEE/CVF conference on computer vision and pattern recognition}, pp.\  11976--11986, 2022.

\bibitem[Millière(2022)]{millire2022adversarial}
Raphaël Millière.
\newblock Adversarial attacks on image generation with made-up words, 2022.

\bibitem[Minderer et~al.(2021)Minderer, Djolonga, Romijnders, Hubis, Zhai, Houlsby, Tran, and Lucic]{rev_cal}
Matthias Minderer, Josip Djolonga, Rob Romijnders, Frances Hubis, Xiaohua Zhai, Neil Houlsby, Dustin Tran, and Mario Lucic.
\newblock Revisiting the calibration of modern neural networks.
\newblock \emph{Advances in Neural Information Processing Systems}, 34:\penalty0 15682--15694, 2021.

\bibitem[Naeini et~al.(2015)Naeini, Cooper, and Hauskrecht]{ece}
Mahdi~Pakdaman Naeini, Gregory~F. Cooper, and Milos Hauskrecht.
\newblock Obtaining well calibrated probabilities using bayesian binning.
\newblock \emph{Proceedings of the ... AAAI Conference on Artificial Intelligence. AAAI Conference on Artificial Intelligence}, 2015:\penalty0 2901--2907, 2015.

\bibitem[Nichol et~al.(2021)Nichol, Dhariwal, Ramesh, Shyam, Mishkin, McGrew, Sutskever, and Chen]{glide}
Alex Nichol, Prafulla Dhariwal, Aditya Ramesh, Pranav Shyam, Pamela Mishkin, Bob McGrew, Ilya Sutskever, and Mark Chen.
\newblock Glide: Towards photorealistic image generation and editing with text-guided diffusion models.
\newblock \emph{arXiv preprint arXiv:2112.10741}, 2021.

\bibitem[Park et~al.(2021)Park, Azadi, Liu, Darrell, and Rohrbach]{composit}
Dong~Huk Park, Samaneh Azadi, Xihui Liu, Trevor Darrell, and Anna Rohrbach.
\newblock Benchmark for compositional text-to-image synthesis.
\newblock In J.~Vanschoren and S.~Yeung (eds.), \emph{Proceedings of the Neural Information Processing Systems Track on Datasets and Benchmarks}, volume~1. Curran, 2021.
\newblock URL \url{https://datasets-benchmarks-proceedings.neurips.cc/paper_files/paper/2021/file/0a09c8844ba8f0936c20bd791130d6b6-Paper-round1.pdf}.

\bibitem[Paszke et~al.(2019)Paszke, Gross, Massa, Lerer, Bradbury, Chanan, Killeen, Lin, Gimelshein, Antiga, et~al.]{pytorch}
Adam Paszke, Sam Gross, Francisco Massa, Adam Lerer, James Bradbury, Gregory Chanan, Trevor Killeen, Zeming Lin, Natalia Gimelshein, Luca Antiga, et~al.
\newblock Pytorch: An imperative style, high-performance deep learning library.
\newblock \emph{Advances in neural information processing systems}, 32, 2019.

\bibitem[Radford et~al.(2021)Radford, Kim, Hallacy, Ramesh, Goh, Agarwal, Sastry, Askell, Mishkin, Clark, et~al.]{clip}
Alec Radford, Jong~Wook Kim, Chris Hallacy, Aditya Ramesh, Gabriel Goh, Sandhini Agarwal, Girish Sastry, Amanda Askell, Pamela Mishkin, Jack Clark, et~al.
\newblock Learning transferable visual models from natural language supervision.
\newblock In \emph{International Conference on Machine Learning}, pp.\  8748--8763. PMLR, 2021.

\bibitem[Ramesh et~al.(2021)Ramesh, Pavlov, Goh, Gray, Voss, Radford, Chen, and Sutskever]{dalle}
Aditya Ramesh, Mikhail Pavlov, Gabriel Goh, Scott Gray, Chelsea Voss, Alec Radford, Mark Chen, and Ilya Sutskever.
\newblock Zero-shot text-to-image generation.
\newblock In \emph{International Conference on Machine Learning}, pp.\  8821--8831. PMLR, 2021.

\bibitem[Ramesh et~al.(2022)Ramesh, Dhariwal, Nichol, Chu, and Chen]{dalle2}
Aditya Ramesh, Prafulla Dhariwal, Alex Nichol, Casey Chu, and Mark Chen.
\newblock Hierarchical text-conditional image generation with clip latents.
\newblock \emph{arXiv preprint arXiv:2204.06125}, 2022.

\bibitem[Rombach et~al.(2022)Rombach, Blattmann, Lorenz, Esser, and Ommer]{ldm}
Robin Rombach, Andreas Blattmann, Dominik Lorenz, Patrick Esser, and Bj{\"o}rn Ommer.
\newblock High-resolution image synthesis with latent diffusion models.
\newblock In \emph{Proceedings of the IEEE/CVF Conference on Computer Vision and Pattern Recognition}, pp.\  10684--10695, 2022.

\bibitem[Saharia et~al.(2022)Saharia, Chan, Saxena, Li, Whang, Denton, Ghasemipour, Ayan, Mahdavi, Lopes, et~al.]{imagen}
Chitwan Saharia, William Chan, Saurabh Saxena, Lala Li, Jay Whang, Emily Denton, Seyed Kamyar~Seyed Ghasemipour, Burcu~Karagol Ayan, S~Sara Mahdavi, Rapha~Gontijo Lopes, et~al.
\newblock Photorealistic text-to-image diffusion models with deep language understanding.
\newblock \emph{arXiv preprint arXiv:2205.11487}, 2022.

\bibitem[Salimans \& Ho(2022)Salimans and Ho]{progressive_sampling}
Tim Salimans and Jonathan Ho.
\newblock Progressive distillation for fast sampling of diffusion models.
\newblock In \emph{International Conference on Learning Representations}, 2022.
\newblock URL \url{https://openreview.net/forum?id=TIdIXIpzhoI}.

\bibitem[Salimans et~al.(2016{\natexlab{a}})Salimans, Goodfellow, Zaremba, Cheung, Radford, Chen, and Chen]{inceptionscore}
Tim Salimans, Ian Goodfellow, Wojciech Zaremba, Vicki Cheung, Alec Radford, Xi~Chen, and Xi~Chen.
\newblock Improved techniques for training gans.
\newblock In D.~Lee, M.~Sugiyama, U.~Luxburg, I.~Guyon, and R.~Garnett (eds.), \emph{Advances in Neural Information Processing Systems}, volume~29. Curran Associates, Inc., 2016{\natexlab{a}}.
\newblock URL \url{https://proceedings.neurips.cc/paper_files/paper/2016/file/8a3363abe792db2d8761d6403605aeb7-Paper.pdf}.

\bibitem[Salimans et~al.(2016{\natexlab{b}})Salimans, Goodfellow, Zaremba, Cheung, Radford, Chen, and Chen]{is_cite}
Tim Salimans, Ian Goodfellow, Wojciech Zaremba, Vicki Cheung, Alec Radford, Xi~Chen, and Xi~Chen.
\newblock Improved techniques for training gans.
\newblock In D.~Lee, M.~Sugiyama, U.~Luxburg, I.~Guyon, and R.~Garnett (eds.), \emph{Advances in Neural Information Processing Systems}, volume~29. Curran Associates, Inc., 2016{\natexlab{b}}.
\newblock URL \url{https://proceedings.neurips.cc/paper_files/paper/2016/file/8a3363abe792db2d8761d6403605aeb7-Paper.pdf}.

\bibitem[Schuhmann et~al.(2021)Schuhmann, Vencu, Beaumont, Kaczmarczyk, Mullis, Katta, Coombes, Jitsev, and Komatsuzaki]{laion400m}
Christoph Schuhmann, Richard Vencu, Romain Beaumont, Robert Kaczmarczyk, Clayton Mullis, Aarush Katta, Theo Coombes, Jenia Jitsev, and Aran Komatsuzaki.
\newblock Laion-400m: Open dataset of clip-filtered 400 million image-text pairs.
\newblock \emph{arXiv preprint arXiv:2111.02114}, 2021.

\bibitem[Schuhmann et~al.(2022)Schuhmann, Beaumont, Vencu, Gordon, Wightman, Cherti, Coombes, Katta, Mullis, Wortsman, et~al.]{laion5b}
Christoph Schuhmann, Romain Beaumont, Richard Vencu, Cade Gordon, Ross Wightman, Mehdi Cherti, Theo Coombes, Aarush Katta, Clayton Mullis, Mitchell Wortsman, et~al.
\newblock Laion-5b: An open large-scale dataset for training next generation image-text models.
\newblock \emph{Advances in Neural Information Processing Systems}, 35:\penalty0 25278--25294, 2022.

\bibitem[Sohl-Dickstein et~al.(2015)Sohl-Dickstein, Weiss, Maheswaranathan, and Ganguli]{diffusion-orig}
Jascha Sohl-Dickstein, Eric Weiss, Niru Maheswaranathan, and Surya Ganguli.
\newblock Deep unsupervised learning using nonequilibrium thermodynamics.
\newblock In Francis Bach and David Blei (eds.), \emph{Proceedings of the 32nd International Conference on Machine Learning}, volume~37 of \emph{Proceedings of Machine Learning Research}, pp.\  2256--2265, Lille, France, 07--09 Jul 2015. PMLR.
\newblock URL \url{https://proceedings.mlr.press/v37/sohl-dickstein15.html}.

\bibitem[Struppek et~al.(2022)Struppek, Hintersdorf, Friedrich, Brack, Schramowski, and Kersting]{struppek2022exploiting}
Lukas Struppek, Dominik Hintersdorf, Felix Friedrich, Manuel Brack, Patrick Schramowski, and Kristian Kersting.
\newblock Exploiting cultural biases via homoglyphs in text-to-image synthesis, 2022.

\bibitem[{TorchVision~maintainers and contributors}(2016)]{torchvision2016}
{TorchVision~maintainers and contributors}.
\newblock Torchvision: Pytorch's computer vision library.
\newblock \url{https://github.com/pytorch/vision}, 2016.

\bibitem[von Platen et~al.(2022)von Platen, Patil, Lozhkov, Cuenca, Lambert, Rasul, Davaadorj, and Wolf]{diffusers}
Patrick von Platen, Suraj Patil, Anton Lozhkov, Pedro Cuenca, Nathan Lambert, Kashif Rasul, Mishig Davaadorj, and Thomas Wolf.
\newblock Diffusers: State-of-the-art diffusion models.
\newblock \url{https://github.com/huggingface/diffusers}, 2022.

\end{thebibliography}

\newpage
\appendix
\onecolumn

\section{Classifier choice}
\label{appendix:classifier_comparison}

The reliance on a pretrained ImageNet classifier is central to our approach, and therefore we investigate how different options for choosing this classifier impact our results.
Specifically, we compute ISP and SCS with three different classifiers: ViT-B/16~\citep{vit}, ConvNeXt-B~\citep{convnext} and ResNet-50~\citep{resnet}. 
The results are displayed in \Cref{tab:full_metrics}: importantly, the values of synset metrics have significant pairwise rank correlations for each specific model (see \Cref{tab:synset_clf_corr}). 
We conclude that while the exact values of ISP and SCS can differ significantly, all classifiers rank the models (along with synsets within one model) in a similar way.

\begin{table}[h]
    \centering
    \caption{Comparison of metric values for different classifiers.}
    \begin{tabular}{lcccccc}
        \toprule
        \multirow{2}{*}{Model} & \multicolumn{3}{c}{In-Subtree Probability $\uparrow$} & \multicolumn{3}{c}{Subtree Coverage Score $\uparrow$}\\ \cmidrule(l){2-4}\cmidrule(l){5-7}
         & ViT-B/16 & ConvNeXt-B & ResNet-50 & ViT-B/16 & ConvNeXt-B & ResNet-50\\ \midrule
         GLIDE & 0.221 & 0.188 & 0.220 & 0.198 & 0.180 & 0.243 \\
         LDM & 0.218 & 0.190 & 0.218 & 0.180 & 0.161 & 0.218 \\
         SD 1.4 & 0.329 & 0.277 & 0.349 & \textbf{0.258} & \textbf{0.221} & \textbf{0.272} \\
         SD 2.0 & 0.296 & 0.254 & 0.307 & 0.232 & 0.205 & 0.259 \\
         unCLIP & \textbf{0.351} & \textbf{0.299} & \textbf{0.363} & 0.190 & 0.157 & 0.211 \\
         \bottomrule
    \end{tabular}
    \label{tab:full_metrics}
\end{table}

\begin{minipage}{0.41\textwidth}
    \centering
    \captionof{table}{Average pairwise Spearman rank correlation between synset metrics for three classifiers. All results are statistically significant (p $<$ 0.05).}
    \begin{tabular}{lcc}
        \toprule
        Model & ISP & SCS \\
        \midrule
        GLIDE & 0.98 & 0.89 \\
        LDM & 0.97 & 0.88 \\
        SD 1.4 & 0.98 & 0.91 \\
        SD 2.0 & 0.98 & 0.91 \\
        unCLIP & 0.97 & 0.89 \\
        \bottomrule
    \end{tabular}
    \label{tab:synset_clf_corr}
\end{minipage}%
\begin{minipage}{0.59\textwidth}
    \centering
    \includegraphics{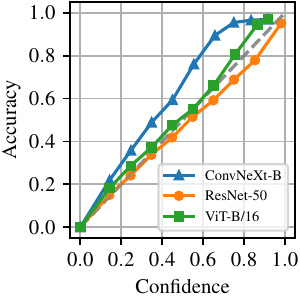}
    \captionof{figure}{Calibration curves on the ImageNet validation set.}
    \label{fig:calibration_curves}
\end{minipage}
\vspace{1em}

To select the best classifier, we compute the expected calibration error (ECE, \citealp{ece}) using 100 bins (following the protocol of \citealp{rev_cal}) and the accuracy on the ImageNet validation set for the three candidate models.
We report the results in \cref{tab:clf_metrics}; in addition, we plot the calibration curves of all models in \cref{fig:calibration_curves}.
Notably, while ConvNeXt-B has the highest accuracy, it is the most miscalibrated model, and therefore we use ViT-B/16 as the classifier for our metrics.

\begin{table}
    \centering
    \caption{Calibration and accuracy metrics on the ImageNet validation set for evaluated classifiers.}
    \begin{tabular}{lcc}
        \toprule
        Classifier & ECE $\downarrow$ & Accuracy $\uparrow$ \\
        \midrule
        ViT-B/16 & \textbf{0.035} & 0.81 \\
        ConvNeXt-B & 0.133 & \textbf{0.84} \\
        ResNet-50 & 0.036 & 0.76 \\
        \bottomrule
    \end{tabular}
    \label{tab:clf_metrics}
\end{table}

\pagebreak
\section{Metric stability}
\label{appendix:metric_stability}

\begin{figure}[h]
    \centering
    \includegraphics[width=0.95\textwidth]{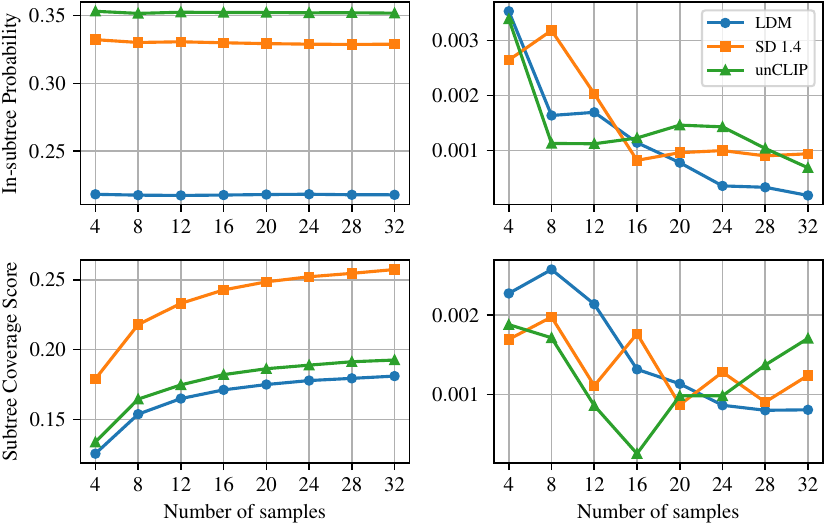}
    \caption{Metric average values (left) and standard deviations (right) as a function of the number of generated samples in each synset.}
    \label{fig:num_samples_stability}
\end{figure}

Here we investigate how changing the number of generated samples per synset affects the final metrics.
To conduct this investigation, we execute four separate runs while varying the number of samples from 4 to 32. 
For each number of samples, we measure the average metric values across runs, as well as their standard deviations.
\cref{fig:num_samples_stability} shows the results of this experiment: we find that both metrics are stable across the analyzed setups with standard deviation rarely exceeding 1\% of the average value. 
We also note that the Subtree Coverage Score increases with the number of samples, which is expected for a diversity measure. 

Furthermore, we analyze how different generated samples impact the ranking of synset ISP and SCS for one model by measuring the Spearman rank correlation between per-synset metrics across four random seeds. 
Our results are available in \Cref{tab:synset_seeds}: we discover that different runs have high pairwise correlations (exceeding $0.97$ on average for ISP and $0.94$ for SCS), concluding that the metrics we design are also stable on a per-synset level.

\begin{table}[h]
    \centering
    \caption{Average pairwise Spearman correlations between synset metrics. All results are statistically significant with p $<$ 0.05.}
    \begin{tabular}{lcc}
        \toprule
        Model & In-Subtree Probability & Subtree Coverage Score \\
        \midrule
        GLIDE & 0.975 & 0.955 \\
        Latent Diffusion & 0.980 & 0.943 \\
        Stable Diffusion 1.4 & 0.983 & 0.947 \\
        Stable Diffusion 2.0 & 0.984 & 0.954 \\
        unCLIP & 0.989 & 0.944 \\
        \bottomrule
    \end{tabular}
    \label{tab:synset_seeds}
\end{table}

\pagebreak
\section{Human evaluation details}
\label{appendix:human_eval}

Our evaluations were conducted on samples from the following 20 synsets: \textit{frog, clock, oven, monkey, knife, wolf, pan, boat, wheel, shark, whale, fruit, turtle, hat, vegetable, pot, flower, duck, chair, spider}. 
The synsets were chosen randomly among those with a distance to the closest leaf node no greater than 2: this was done to eliminate overly abstract concepts for ease of interpretation by crowd workers. 
We manually discarded words with different possible meanings, such as ``rail''.

\begin{figure}
    \centering
    \begin{subfigure}[b]{\textwidth}
    \includegraphics[width=\textwidth]{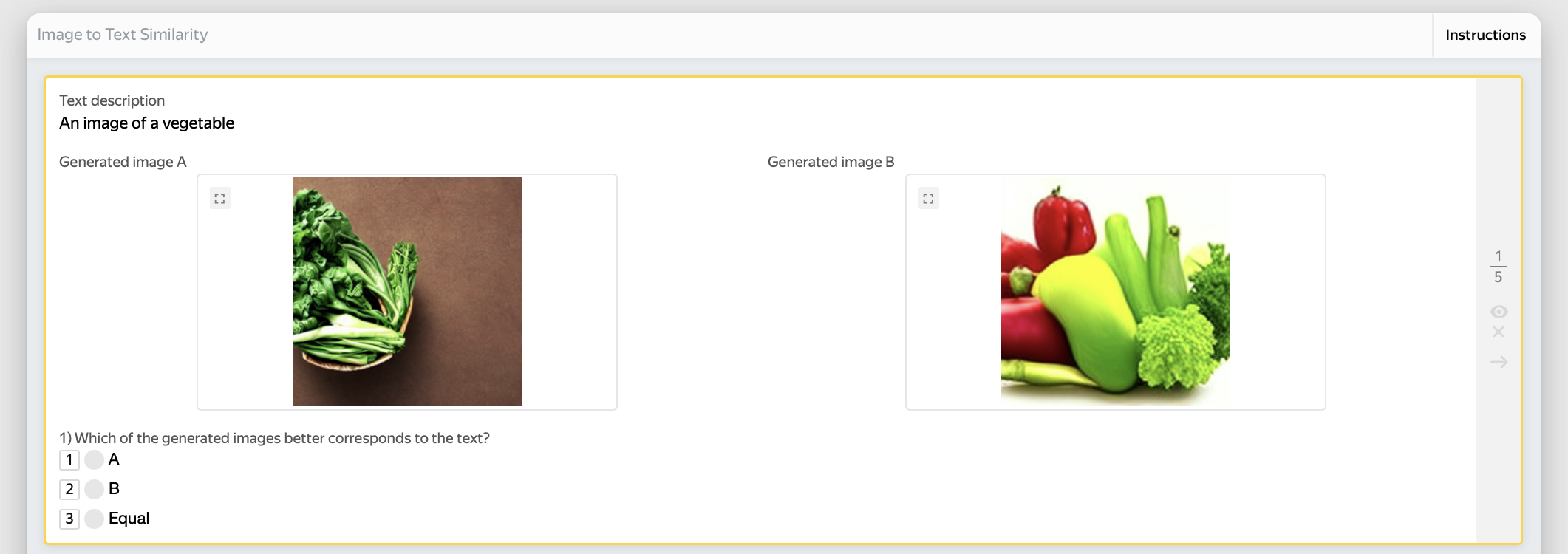}
    \caption{Caption similarity task.}
    \end{subfigure}
    \vspace{18pt}

    \begin{subfigure}[b]{\textwidth}
    \includegraphics[width=\textwidth]{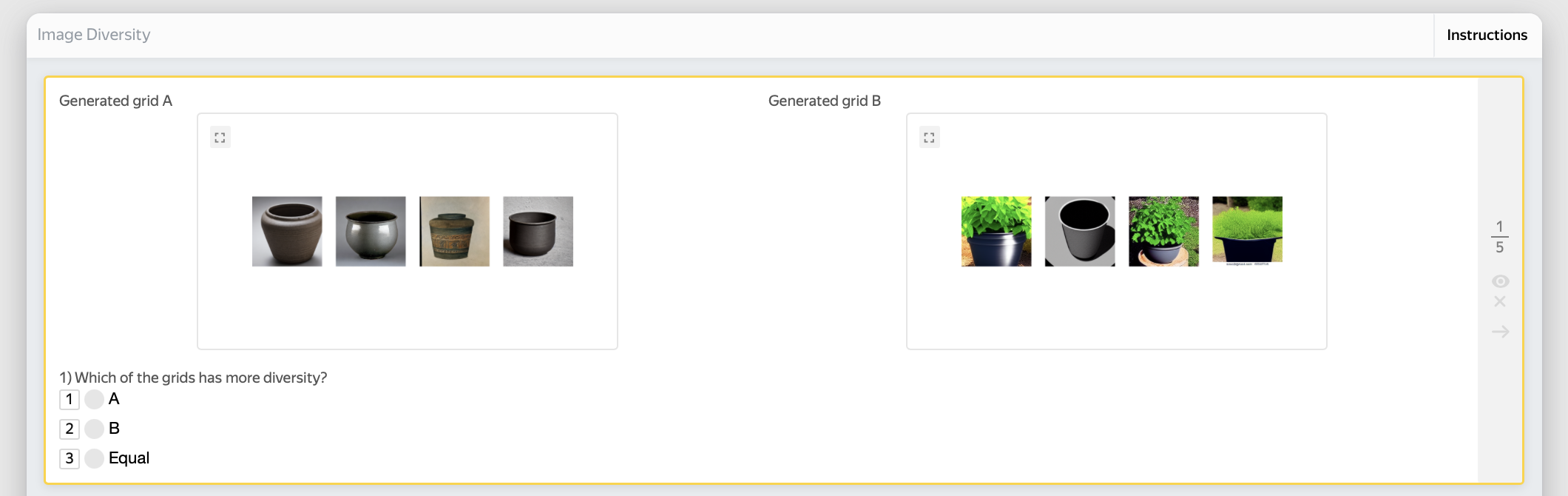}
    \caption{Diversity task.}
    \end{subfigure}
    \caption{Screenshots from the annotation interface.}
    \label{fig:eval_samples}
\end{figure}

We provide task descriptions in Figures~\ref{fig:desc_caption} and~\ref{ref:desc_diversity}, and the evaluation interface is shown in Figure~\ref{fig:eval_samples}. The participants were paid \$0.10 per task, which exceeds the hourly minimum wage in their geographical regions. 
We required that participants complete 5 manually labeled training tasks and achieve an accuracy of more than 60\% on them before starting the evaluation procedure.

\begin{figure}[h]
    \noindent\rule{\textwidth}{1pt}\\

    Two neural networks tried to generate an image of an object given the text caption.
    Please help us understand which image better matches the object given in the text.\\
    
    \textbf{How to answer the question:}
    
    For all comparisons, we provide a text description from which these images were created. Text description contains a reference to some object (e.g. ``An image of a cat''). To answer the question, we suggest using the following algorithm. For each generated image:
    \begin{itemize}
        \item First, read the text description (e.g., ``An image of a cat'').
        \item If no images correspond to the object, select option ``equal''.
        \item If only one image corresponds to the object and another one does not: select the image that corresponds to the text. 
        \item If both images correspond to the text: it is up to you whether to select ``equal'' or to choose the one that corresponds to the text more precisely.
    \end{itemize}
    \noindent\rule{\textwidth}{1pt}
    \vspace{-12pt}
    \caption{Text to caption similarity task description.}
    \label{fig:desc_caption}
    \vspace{-12pt}
\end{figure}

\begin{figure}[h]
    \noindent\rule{\textwidth}{1pt}\\

    Two neural networks tried to generate an image of an object given the text caption. We present to you two grids of 4 generated images each.
    Please help us understand which grid of images is more diverse.\\
    
    \textbf{What do we mean by diversity:}
    
    A grid is diverse if it has variation in the generated object. Some examples of variation include:
    \begin{itemize}
    \item Different animal species (e.g., a Persian cat and a sphinx cat).
    \item Different subtypes of an object: (e.g., a race car, a sedan car).
    \item Different colors: (e.g., a black cat and a white cat).
    \item Different positions of the same object (e.g., a running human and a sitting human).
    \item Different details on the same object (e.g., a human wearing glasses and a human wearing a monocle).
    \end{itemize}

    \textbf{How to answer the question:}
    
    To answer the question, we suggest using the following algorithm. For each pair of grids:
    \begin{itemize}
    \item If only one grid has diverse images and the other one has little variation: select the grid that is diverse.
    \item If none of the grids has diverse images, and both of them have little variation: select ``equal''. 
    \item If both images have some level of diversity, it's up to you whether to select ``equal'' or to choose the one that has more diversity.
    \end{itemize}
    \noindent\rule{\textwidth}{1pt}
    \vspace{-12pt}
    \caption{Diversity task description.}
    \label{ref:desc_diversity}
\end{figure}

The tasks were presented in groups of five. We included one control task in each group to filter low-quality annotations. 
For text-to-caption similarity, the control tasks had one regular image and one image generated from a different synset. 
For image diversity, control tasks had one set of regular images and one grid that consisted of four identical images.
Participants who failed two control tasks in a row were banned. 
We also included measures against responses significantly faster than the estimated time for completing a task.

\section{Additional synsets with low In-subtree Probability}
\label{appendix:low_in_random}

\begin{figure}
    \centering
    \includegraphics{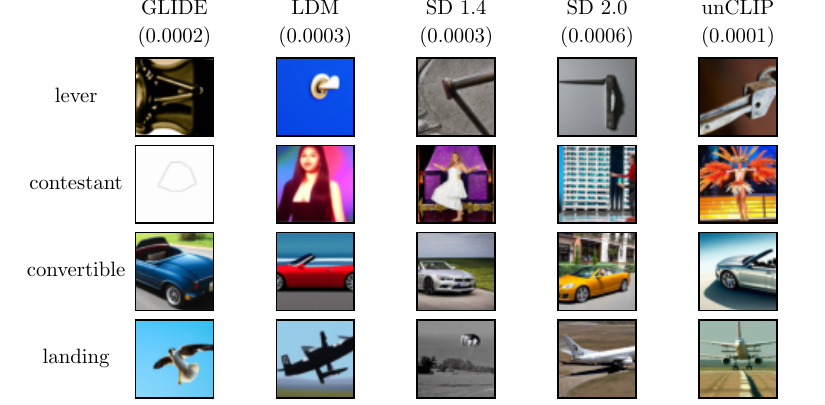}
    \caption{Generated images of randomly selected synsets with low ISP. Average model ISP for these synsets is presented in parentheses.}
    \label{fig:low_in_synsets_random}
\end{figure}

As discussed in \cref{section:low_isp}, our approach allows discovering concepts that are unknown to the model by selecting synsets with low In-Subtree Probability. 
For reproducibility purposes, we also present a random selection of such concepts in \Cref{fig:low_in_synsets_random}. 
These concepts were sampled from the lowest 50 synsets in terms of average ISP across all models used in our study.

\section{Finding concepts with low diversity}
\label{appendix:low_diversity}

\begin{figure}[t]
    \centering
    \includegraphics{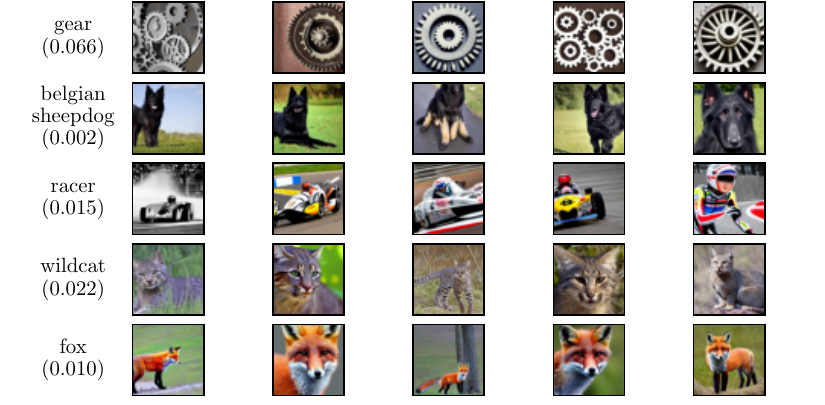}
    \caption{Generated images of randomly selected synsets with low SCS for Stable Diffusion 1.4. Synset SCS is presented in parentheses.}
    \label{fig:low_scs_synsets}
\end{figure}

Similarly to the analysis of ~\cref{section:low_isp}, it is also possible to find concepts that have low diversity for the given model.
We analyze Stable Diffusion 1.4 by selecting random concepts with a low Subtree Coverage Score and displaying them in \ref{fig:low_scs_synsets}. 
Our findings are highly interpretable: for example, ``belgian sheepdog'' has four varieties: ``groenendael'', ``malinois'', ``tervuren'' and ``laekenois'', and only the first two are parts of the ImageNet hierarchy. 
The model only draws the ``groenendael'', which results in a low coverage score.

\pagebreak
\section{Subtree comparison}
\label{appendix:subtree_comp}

Our approach makes it easy to evaluate models on a particular set of concepts by simply averaging the metrics for the synsets corresponding to those concepts. 
The hierarchical nature of the ImageNet tree also simplifies the process of finding large sets of semantically connected words: one can simply take entire hyponym subtrees of concepts of interest. 

We compare models from \cref{section:experiments} on an example set of concept subtrees, reporting the results in \Cref{tab:isp_subtree,tab:scs_subtree}.
Notably, model rankings on these subtrees significantly differ from those given by metrics over the entire hierarchy. 
This difference highlights the advantage of ISP and SCS: we are able to go beyond what a single metric value would give us.
For example, it is possible to focus on a specific group of concepts depending on the target application or when choosing the best-performing model for the target prompt.

\begin{table}[h]
    \centering
    \caption{In-Subtree Probability for different subtrees of the ImageNet hierarchy. The highest value is in bold, the second highest is underlined.}
    \begin{tabular}{lccccc}
        \toprule
        Subset & GLIDE & LDM & Stable Diffusion 1.4 & Stable Diffusion 2.0 & unCLIP\\
        \midrule
        Vessel & 0.282 & 0.497 & 0.512 & \underline{0.578} & \textbf{0.584}\\
        Furniture & 0.267 & 0.333 & \textbf{0.481} & 0.384 & \underline{0.436}\\
        Bird & \textbf{0.470} & 0.271 & \underline{0.462} & 0.420 & 0.425\\
        Clothing & 0.065 & 0.206 & \underline{0.247} & 0.172 & \textbf{0.276}\\
        Lizard & \textbf{0.346} & 0.175 & 0.289 & 0.263 & \underline{0.295}\\
        Fruit & \textbf{0.492} & 0.374 & 0.438 & 0.329 & \underline{0.452}\\
        \midrule
        Full hierarchy & 0.221 & 0.218 & \underline{0.329} & 0.296 & \textbf{0.351}\\
        \bottomrule
    \end{tabular}
    \label{tab:isp_subtree}
\end{table}

\begin{table}
    \centering
    \caption{Subtree Coverage Score for different subtrees of the ImageNet hierarchy. The highest value is in bold, the second highest is underlined.}
    \begin{tabular}{lccccc}
        \toprule
        Subset & GLIDE & LDM & Stable Diffusion 1.4 & Stable Diffusion 2.0 & unCLIP\\
        \midrule
        Vessel & 0.188 & 0.183 & \textbf{0.267} & \underline{0.205} & 0.187\\
        Furniture & \textbf{0.211} & 0.167 & \underline{0.190} & 0.182 & 0.183\\
        Bird & 0.152 & 0.137 & \textbf{0.168} & \underline{0.160} & 0.092\\
        Clothing & 0.090 & \underline{0.160} & \textbf{0.193} & 0.148 & 0.139\\
        Lizard & 0.084 & \underline{0.104} & 0.064 & \textbf{0.119} & 0.086\\
        Fruit & 0.203 & 0.158 & \underline{0.218} & \textbf{0.240} & 0.205\\
        \midrule
        Entire hierarchy & 0.198 & 0.180 & \textbf{0.258} & \underline{0.232} & 0.190\\
        \bottomrule
    \end{tabular}
    \label{tab:scs_subtree}
\end{table}

\section{Relationship with the textual encoder}
\label{appendix:textual_encoder}
As the metric values for models vary across synsets, a natural question is whether the quality for a given concept corresponds to the knowledge about this concept contained in the textual encoder of the model.
To verify this, we conduct a comparison of performance across synsets with the similarity of each synset to its hyponyms, using the values of ISP and SCS for Stable Diffusion 1.4, which uses CLIP ViT-L/14 text encoder for conditioning on its prompts.

More specifically, for each synset from the evaluation set, we obtain the CLIP text encoder embeddings for this synset, as well as the embeddings for all its hyponyms contained in the set of ImageNet classes.
We exclude all other hyponyms to ensure a proper comparison with ISP and SCS.
After that, we compute the average cosine similarity of each synset to its hyponyms and calculate the correlation of these similarities to ISP and SCS across a range of classifier-free guidance values.

\begin{table}[h]
\centering
\caption{Spearman correlation of CLIP hyponym similarities with WordNet-based metrics for Stable Diffusion 1.4. All results are statistically significant (p $<$ 0.05).}
\label{cosine}
\begin{tabular}{lcc}
\toprule Guidance & In-Subtree Probability & Subtree Coverage Score \\ 
\midrule 2.5 & 0.397 & -0.139 \\
5.0 & 0.405 & -0.178 \\
7.5 & 0.400 & -0.186 \\
10.0 & 0.393 & -0.192 \\
\bottomrule
\end{tabular}
\end{table}

The results of this evaluation are available in \cref{cosine}. 
As we can see, the cosine similarity of synsets to their hyponyms significantly correlates with the In-Subtree Probability, which suggests a connection between the knowledge of the hypernymy relationship of the encoder and the performance of the entire model according to this metric.
On the other hand, the Subtree Coverage Score displays a negative correlation, which might be caused by more diverse subtrees with an inaccurate representation of the prompt having higher scores.

\end{document}